\documentclass[12pt, a4paper, notitlepage, oneside]{article}

\usepackage{graphicx}
\usepackage[utf8]{inputenc}
\usepackage{amsmath}
\usepackage{amssymb}
\usepackage{hyperref}
\usepackage{float}

\title{Demonstrating Rosa: the fairness solution for any Data Analytic pipeline}

\author{ Kate Wilkinson  \\ illumr Ltd., London, United Kingdom  \and George \v{C}evora   \\ illumr Ltd., London, United Kingdom \\ \href{mailto:george.cevora@illumr.com}{george.cevora@illumr.com}}

\date{}

\begin{document}
	
	\maketitle
	
	\abstract{ Most datasets of interest to the analytics industry are impacted by various forms of human bias. The outcomes of Data Analytics [DA] or Machine Learning [ML] on such data are therefore prone to replicating the bias. As a result, a large number of biased decision-making systems based on DA/ML have recently attracted attention. 

In this paper we introduce Rosa, a free, web-based tool to easily de-bias datasets with respect to a chosen characteristic. Rosa is based on the principles of Fair Adversarial Networks, developed by illumr Ltd., and can therefore remove interactive, non-linear, and non-binary bias. Rosa is stand-alone pre-processing step / API, meaning it can be used easily with any DA/ML pipeline.

We test the efficacy of Rosa in removing bias from data-driven decision making systems by performing standard DA tasks on five real-world datasets, selected for their relevance to current DA problems, and also their high potential for bias. We use simple ML models to model a characteristic of analytical interest, and compare the level of bias in the model output both with and without Rosa as a pre-processing step. 

We find that in all cases there is a substantial decrease in bias of the data-driven decision making systems when the data is pre-processed with Rosa.}

	\newpage

\label{sec:Intro}

A topic that has recently received much attention both in the media and in academia is the bias in decisions made based on the output of Machine Learning [ML] algorithms \cite{zou2018ai}\cite{o2016weapons}\cite{eubanks2018automating}. This bias is especially problematic when the result is unfair discrimination between groups on the basis of a protected characteristic. In UK Law, protected characteristics are defined as age, disability, gender reassignment, marriage and civil partnership, pregnancy and maternity, race, religion or belief, sex and sexual orientation. The Equality Act, 2010 \cite{act2010c}, made it illegal to discriminate based on any of these characteristics. 

Discrimination in the output of ML algorithms most often stems from biased human judgements made in the past. These biased judgements can make up the dataset which an algorithm is trained on, or they can shape the composition of the dataset so that it does not represent the true population. If an ML algorithm is trained to predict biased human judgements, then the bias will be replicated in decisions made based on the output of the algorithm.

One widely publicised example occurred in 2015, when Amazon revealed that its experimental recruitment algorithm was heavily biased against women \cite{amazonReuters}. The dataset used to train the algorithm comprised successful resumes submitted to the company over the last 10 years. The technology industry is historically and currently male-dominated, and therefore the proportion of female resumes in the dataset was far less than the true proportion of female applicants that would be suitable for the job today. It is understandable that an algorithm trained to assess the suitability of applicants would find a link between successful applicants and gender. If decisions are made based on the assessments of this algorithm, then an unfair proportion of female applicants will be hired, propagating the bias further.

Almost every dataset has some degree of bias, and not using biased data at all is certainly worse than making decisions without any data. This calls for methods of debiasing data with respect to a given characteristic, so that we can still use potentially biased datasets without fear of creating algorithms that discriminate unfairly.

Unfortunately, removing bias from a dataset or algorithm is not straightforward. Simply removing the column that contains information about the protected characteristic does not necessarily remove information about that characteristic from the dataset. Datasets often contain variables which are proxies for other variables. For example, postcode may be a proxy for race, or education may be a proxy for gender. As discussed in \v{C}evora 2019 \cite{fans2018cevora}, the most common
methods of bias removal do not work particularly well, are based on subjective notions of fairness and/or are very difficult to implement.

Fair Adversarial Networks [FANs] are an alternative technique for removing multivariate, non-binary and/or non-linear bias from a dataset, whilst being trivial to implement \cite{fans2018cevora}. The FAN methodology encodes a novel, biased dataset which can subsequently be used with a range of analytical tools, free of bias. FANs therefore directly tackle the problem of biased datasets in ML.

In December 2019, illumr Ltd. released a tool called Rosa, which uses the principles of FANs to debias novel datasets using a web-based interface. A free, demo version of Rosa is available online (at \href{https://illumr.com/rosa}{illumr's website}), where any user can upload a small dataset and debias it with respect to a chosen characteristic. In particular, Rosa aims to be accessible to data analysts who may not have much experience with data preparation or advanced Machine Learning techniques.

In this paper, we demonstrate the effect of Rosa on 5, real-world datasets. The datasets are selected for their high potential to be used in real-world decision making, and also their high potential for inherent bias.

To test the effectiveness of Rosa, we perform a simple data analytical task using each dataset, and examine the degree of bias in the output. We then repeat the task using datasets which have been processed by Rosa, and investigate whether the bias has decreased. We find significant/total reductions in bias on all datasets after processing by Rosa. The models that are chosen for the analytical tasks are basic, out-of-the box models, in order to imitate the scenario in which an inexperienced analyst may wish to use Rosa.

\newpage
\section{Data Analytic Examples}
\subsection{Criminal Recidivism}
\label{sec:compas_intro}
Criminal sanctions in most countries do not only depend solely on the criminal act that has been committed, but also on personal circumstances (e.g. having children) and the level of threat the individual poses to society. Of a particular interest to us is estimation of the likelihood of recidivism for a criminal defendant, as it is becoming increasingly common to delegate this task to automated decision-making systems. These systems can have a huge impact on people's lives, as individuals deemed unlikely to recidivate may be released on bail or assigned lower sentences than those deemed otherwise.

\href{https://www.equivant.com/northpointe-suite/}{Correctional Offender Management Profiling for Alternative Sanctions} [COMPAS] is a technology used by the Department of Justice of the United States [DoJ] to assess the likelihood of recidivism of criminal defendants and guide bail decisions. Larson and colleagues \cite{larson2016we} found significant racial bias in these assessments, with African-Americans being assigned significantly higher recidivism risk scores, independent of whether they recidivated or not. This means COMPAS is much more likely to wrongly label African-Americans as potential recidivists than White Americans. Conversely White Americans are much more likely to be wrongly labelled as non-recidivists.

The DoJ unfortunately defends this pattern of discrimination as being fair \cite{flores2016false}. Their argument relies on the fact that more African-Americans recidivate according to DoJ statistics. Shockingly, Flores and colleagues, writing in defense of the DoJ, fail to acknowledge the significant evidence for racial bias within the criminal justice system itself, with Black populations suffering from higher rates of stop, search, and arrest, despite research to suggest that their crime rate is relatively similar to other populations \cite{piquero2008assessing,dabney2006impact,bunting2013war,dabney2004actually,schreer2009shopping,shallice1990black}. As a result, crime datasets often imply a much higher rate of criminality for Black populations than is likely to be the case.

\subsubsection{Data and Methods}
In this section we use a dataset complied by ProPublica that includes the criminal profiles of 11,757 defendants from Broward
County, Florida, along with an indication of whether they recidivated during  a two-year follow-up study. The dataset is freely available to download at \href{https://github.com/propublica/compas-analysis}{Github}.  Before starting the analysis we discarded non-categorical or numerical data, and data relating to the COMPAS risk-assessment scores allocated to the offenders. We kept data from only two race categories, \emph{Caucasian} and \emph{African-American}. 

To replicate a recidivism model such as COMPAS we used a basic logistic regression model from Python's sklearn package to predict likelihood of recidivism while excluding the race variable from the data, and looked for bias in the predictions of our model. We then processed the data with Rosa and repeated the exact same modelling, checking to see whether the bias had reduced.

\subsubsection{Results}
\begin{figure}[H]
	\centering
	\begin{tabular}{cc}
		\includegraphics[width=7cm]{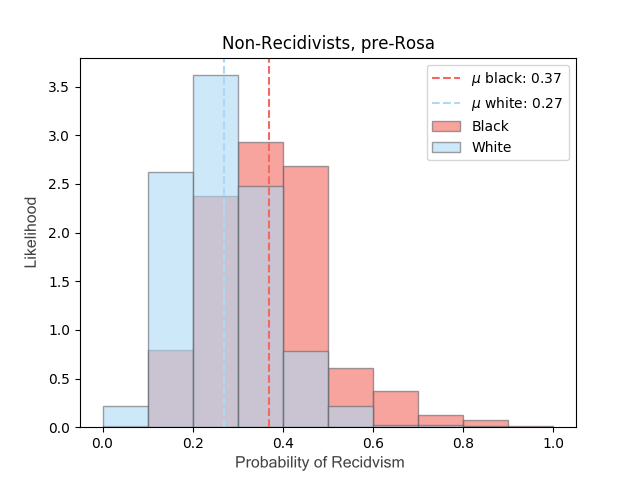} & \includegraphics[width=7cm]{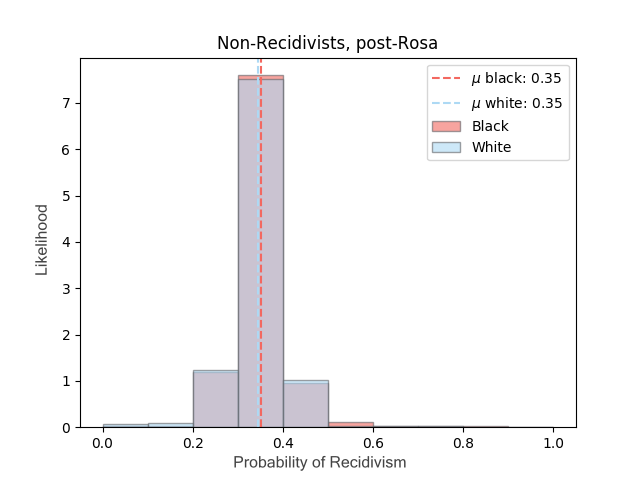} \\
		\includegraphics[width=7cm]{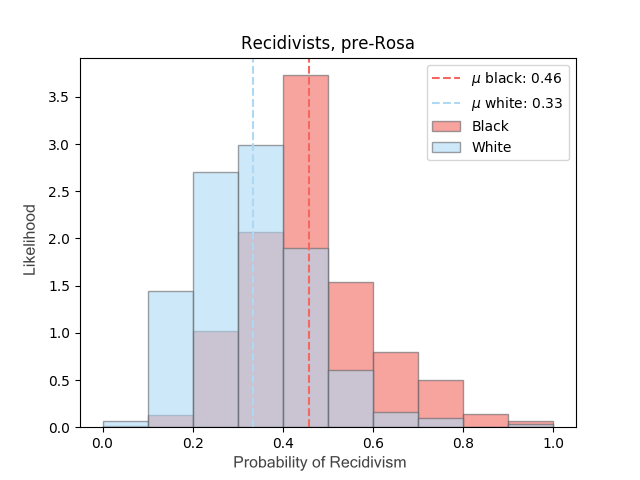} & \includegraphics[width=7cm]{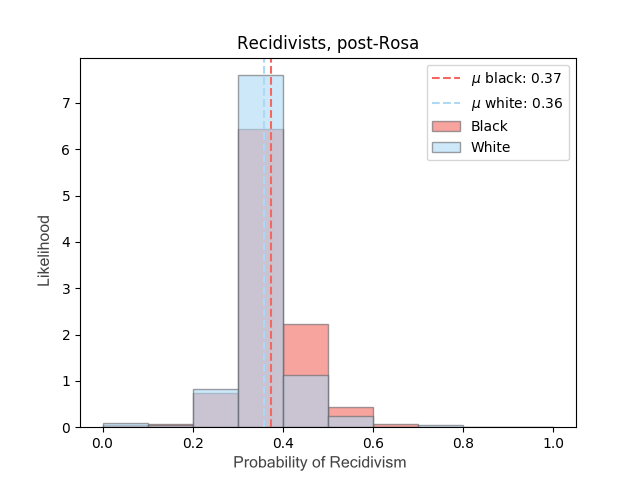} \\
	\end{tabular}
	\caption{Without data pre-processing using Rosa (left-hand plots), African-Americans are labelled as having a higher probability of recidivism than their Caucasian counterparts by our model, independent of whether they recidivated or not. Using Rosa as a pre-processing step (right-hand plot), this difference has been almost entirely removed. Note that without considering the estimates for recidivists and non-recidivists separately, any difference in the score distribution for African-Americans and Caucasians could be explained by a different rate of recidivism between the racial categories, as opposed to an unfair pattern of misclassification.}
	\label{fig:compas_hist}
\end{figure}

As shown in Figures \ref{fig:compas_hist} and \ref{fig:compas_table} the distribution of model estimates is clearly discriminatory against African-Americans when Rosa is not used. This bias essentially disappeared when the data was pre-processed using Rosa.

\begin{figure}[H]
	\vspace{20pt}
	    \centering
    \begin{tabular}{r||cc|cc}
         & \multicolumn{2}{c|}{\textbf{Original data}}&\multicolumn{2}{c}{\textbf{Rosa pre-processed}} \\
         &non-recid &recid &non-recid &recid\\
         \hline
         \hline
         $\mu$ African-American& 0.37 & 0. 46 & 0.35 & 0.37\\
         $\mu$ Caucasian& 0.27 & 0.33 & 0.35 & 0.36\\
         $\mu$ Diff & 0.10 & 0.13 & 0.00 & 0.01 \\
         \hline
         $\sigma$ African-American & 0.13 & 0.14 & 0.06 & 0.06 \\
         $\sigma$ Caucasian &  0.11 & 0.13 & 0.05 & 0.06 \\
         $\sigma$ Average &  0.12 & 0.14 & 0.06 & 0.06 \\
         \hline
         $\mu$ Diff / $\sigma$ Average & 0.85 & 0.92 & 0.00 & 0.17\\
         
    \end{tabular}
    
	\caption{Caucasian individuals receive significantly lower recidivism likelihood estimates from our recidivism model when Rosa was not used than their African-American counterparts, irrespective of whether they later recidivate or not. The amount of bias in the model has been quantified by dividing the difference in mean estimate for African-Americans and Caucasians by the average standard deviation of the two distributions. The higher the value the greater the bias. 
	Using Rosa has significantly decreased the level of bias in the estimates both for recidivists and non-recidivists. }
	\label{fig:compas_table}
\end{figure}

The accuracy of the regression model prior to debiasing by Rosa was 67\%, and post-debiasing it was 63\%. This drop, while significant is likely due to the bias inherent in the dataset - which was also used to evaluate accuracy. Unfortunately the real-world nature of the examples presented in this paper do not allow unbiased evaluation.

\subsubsection{Discussion}
We have performed a simple DA task to model the chance of recidivism of a criminal defendant. Similar systems are used across USA to determine a defendant's suitability for alternative sanctions and can therefore have a significant impact on the individual's life. Without correcting for bias our model discriminated against African-Americans, who were always considered of higher recidivism risk whether they recidivated or not. These results are in line with an analysis of COMPAS - a system developed by Equivant and used by DoJ that has been demonstrated to perform similar discrimination \cite{larson2016we}.

Replicating the exact same DA pipeline that resulted in a discriminatory model, but applying Rosa as a data pre-processing step, resulted in model that did not discriminate.

It is hardly surprising that the initial model was biased against African-Americans. The dataset was likely affected by the issues discussed in Section \ref{sec:compas_intro}, such as a heightened rate of stop and search for African-Americans compared to Caucasians, leading to an over-representation of African-Americans (and black people in general) in criminal datasets relative to their rate of criminality.

For example, arrests for marijuana possession are 3.7 times higher for the black population in the US
than the white population, despite similar reported levels of usage \cite{bunting2013war}. For the COMPAS dataset, `Marijuana' was one of the most frequently occurring words in the text describing the charge for each defendant. 
Dealing with such systematic bias in criminal datasets is essential as data-driven decision making becomes more widespread. Here we have demonstrated that Rosa is a suitable tool for such a task.

\newpage
\subsection{Absenteeism at work}
\label{sec:absenteeism}
DA has an ever-increasing role in Human Resource Management [HR], with employers aiming to only hire/retain the highest performing employees. Absenteeism is an important aspect of HR for many organisations and therefore its estimation is an interesting feature that can be used to screen applicants during the hiring process. While employers should be considering absenteeism during the hiring process, it is known to be related to age \cite{martocchio1989age} which is a protected characteristic. Achieving lower absenteeism in a workforce by age-skewed hiring would therefore be illegal.

\subsubsection{Data and Methods}

We have performed a simple analysis to model absenteeism in the \emph{Absenteeism at work} dataset available from the \href{https://archive.ics.uci.edu/ml/datasets/Absenteeism+at+work}{UCI Machine Learning Repository}. This dataset holds personal information on a set of employees from a courier company in Brazil, along with their total absenteeism time over three years. 

Before commencing with the modelling exercise we removed data on the type/season of absence, and replaced the `Absenteeism Time in Hours' variable  with a derived feature: `Absenteeism in Upper Quartile', which indicates whether an employee has absence in the upper quartile of all employee absences. We also replaced the age (in years) with a categorical variable, indicating whether employees were under 35, 35 to 45 or over 45. This simplified the identification of bias as we could compare the model estimates of absenteeism directly between the three age groups, rather than in a continuous fashion.

We used the Logistic Regression model from Python's sklearn package to predict whether employees had absenteeism in the upper quartile of all the employees in the dataset while excluding the age variable from the data, and analysed for bias with respect to age. We then used the same model to predict absenteeism with data that had been debiased by Rosa, and checked whether the bias has reduced.

\subsubsection{Results}

\begin{figure}[H]
	\centering
	\begin{tabular}{cc}
		\includegraphics[width=7cm]{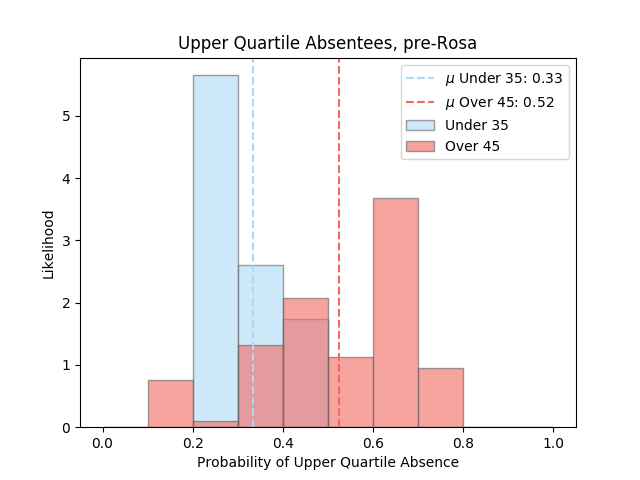} & \includegraphics[width=7cm]{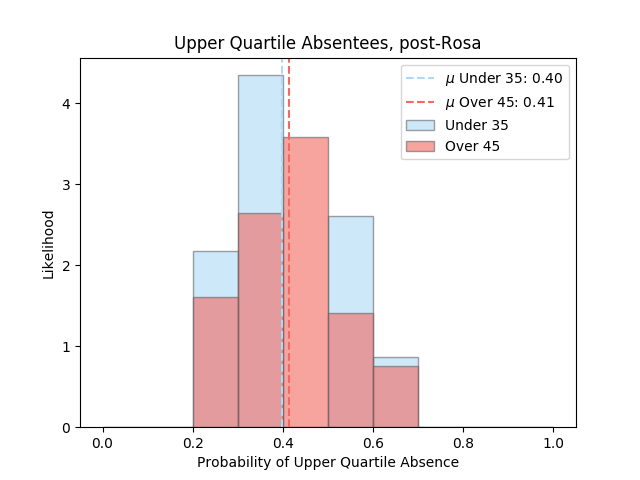} \\
		\includegraphics[width=7cm]{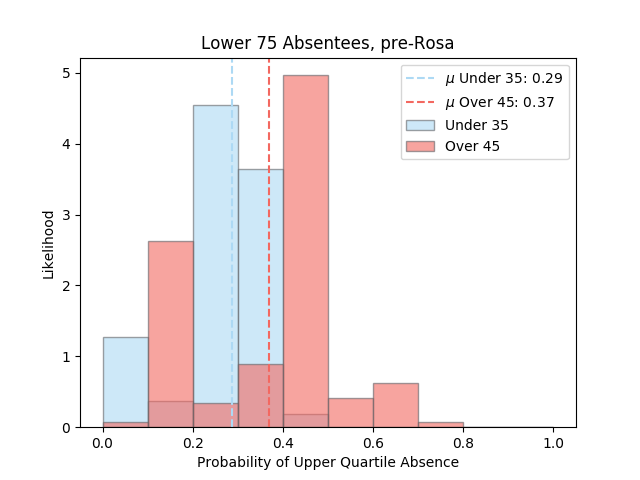} & \includegraphics[width=7cm]{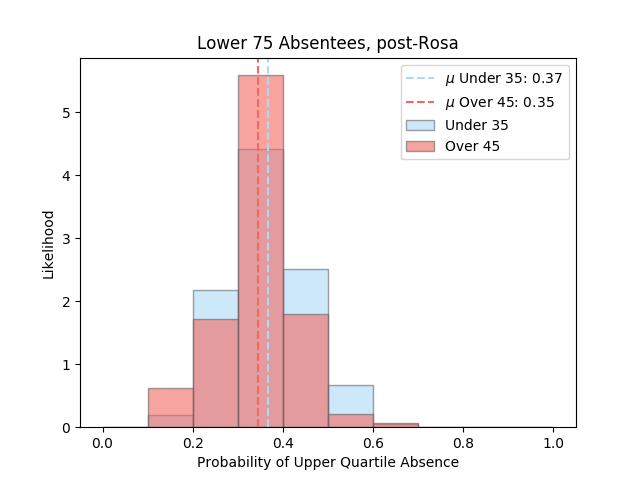} \\
	\end{tabular}
	\caption{For both the upper quartile and lower three-quartile absentees, those under 35 received lower estimates of absenteeism than those over 45. The right-hand plots show the model estimates using data pre-processed by Rosa, where most of the difference in model estimates for the over 45 and under 35 groups has been removed. In particular, before Rosa was applied to the dataset, the older lower three-quartile absentees received higher estimates of absenteeism than the younger upper quartile absentees, indicating a high degree of bias with respect to age. This pattern disappeared when Rosa was used  to pre-process the data.}
	\label{fig:absenteeism_hist}
\end{figure}

\begin{figure}[H]
	\vspace{20pt}
	    \centering
    \begin{tabular}{r||cc|cc}
         & \multicolumn{2}{c|}{\textbf{Original data}}&\multicolumn{2}{c}{\textbf{Rosa pre-processed}} \\
         &upper quart &lower quarts &upper quart &lower quarts\\
         \hline
         \hline
         $\mu$ Under 35 & 0.33 & 0.29 & 0.40 & 0.37\\
         $\mu$ Over 45& 0.52 & 0.37 & 0.41 & 0.35 \\
         $\mu$ Diff & 0.19 & 0.08 & 0.01 & 0.02 \\
         \hline
         $\sigma$ Under 35 & 0.05 & 0.10 & 0.17 & 0.15 \\
         $\sigma$ Over 45 & 0.16 & 0.14 & 0.11 & 0.08 \\
         $\sigma$ Average & 0.11 & 0.12 & 0.12 & 0.12 \\
         \hline
         $\mu$ Diff / $\sigma$ Average & 1.81 & 0.69 & 0.13 & 0.20 \\
    \end{tabular}
    
	\caption{Younger individuals receive significantly lower estimates of absenteeism than their older counterparts when we investigate the upper and lower three-quartiles of absentees separately in a model without any compensation of age related bias. The amount of bias in the model has been quantified by dividing the difference in mean estimate for those over 45 and those under  35 by the average standard deviation of the two distributions. The higher the value the greater the bias. Using Rosa has significantly decreased the level of bias in the estimates both for those with upper quartile absenteeism and those without.}
	\label{fig:absenteeism_table}
\end{figure} 

There was a decrease in prediction accuracy of the model from 72\% to 65\%, although the unbiased prediction accuracy cannot be directly compared with the prediction accuracy of a biased model. This is because predicting certain outcomes with and without bias are fundamentally different tasks, and therefore success on these separate tasks cannot be compared directly.

\subsubsection{Discussion} 

We demonstrated that a simple model to predict employee absenteeism shows significant bias with respect to age. This is concerning as it is becoming increasingly common for automated assessments to form a part of hiring processes, and predicting absenteeism is of much interest to employers. While it is acceptable to discriminate based on likely absenteeism alone, we have seen that this correlates with age in a simple model, which would likely result in age-based discrimination in hiring, which is illegal. 

A particular pattern of discrimination appeared in the analysis of the original dataset: the older lower three-quartile absentees received higher estimates of absenteeism than the younger upper-quartile absentees, indicating a large degree of discrimination with respect to age. This pattern  disappeared when Rosa was used to pre-process the data.

Using the exact same DA pipeline to predict absenteeism after pre-processing by Rosa, the model estimates showed near zero bias, meaning that the dataset would be safe to use in an automated hiring procedure without risk of age discrimination.

\subsection{Heart Disease}
\label{sec:heart}
Coronary heart disease [CHD] is the leading cause of death in women \cite{abubakar2015global}. At certain ages, CHD mortality rates are up to 5x higher in men than women \cite{Botse000298}. However, this risk is highly age dependent and means that over a lifetime, men and women have a relatively similar risk. Despite this, the higher risk for men during middle-age leads to the common misconception that CHD is a `men's disease', in turn leading to a lack of research and data collected on CHD in women.

Although the major risk factors for CHD in healthy women are the same as those identified for healthy men in epidemiological studies, the relative strength of certain risk factors may depend on gender \cite{manson1992primary}\cite{vittinghoff2003risk}. This means that a model which is trained to predict the risk of heart disease using data mostly from men may perform less well on women. This can also lead to the misdiagnosis of heart disease in women, as symptoms that indicate the presence of heart disease in women are often considered atypical \cite{mcsweeney2005s}.

Misdiagnosis is a serious issue, as the longer a patient goes without appropriate treatment, the greater the risk of mortality. A study by the University of Leeds found that women who had a final diagnosis of a STEMI-type heart attack had a 59 per cent greater chance of a misdiagnosis compared with men \cite{wu2018editor}. 

\subsubsection{Data and Methods}

We used a simple DA model to predict the presence of heart disease from a dataset of various blood measurements from patients in Cleveland, Ohio. The \emph{Heart Disease Dataset} is available at \href{https://www.kaggle.com/ronitf/heart-disease-uci}{Kaggle.} The original dataset has a scale for type of heart disease, with 0 indicating the absence of heart disease and 1 to 3 indicating some degree of heart disease. To simplify modelling, we converted these categories into a binary variable indicating whether a patient has any degree of heart disease or not.

We used the Logistic Regression model from Python's sklearn package to predict the presence of heart disease for individuals in the dataset while excluding the gender variable from the data, and checked for bias with respect to gender in the model estimates. We then debiased the data with respect to gender, and repeated the exact same analytical process to see whether the bias had decreased.

\subsubsection{Results}
\begin{figure}[H]
	\centering
	\begin{tabular}{ll}
		\includegraphics[width=7cm]{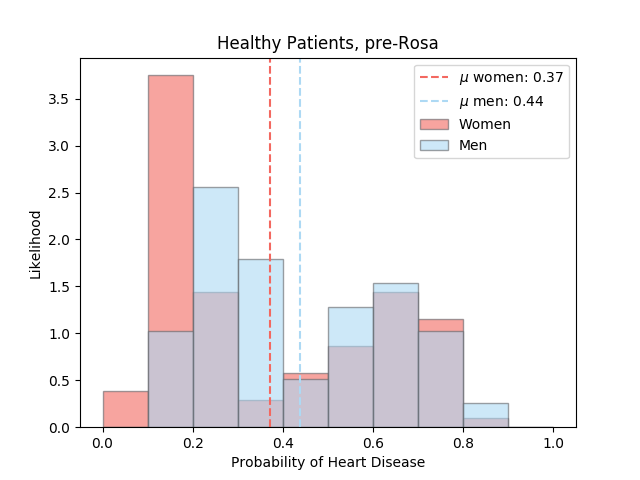} & \includegraphics[width=7cm]{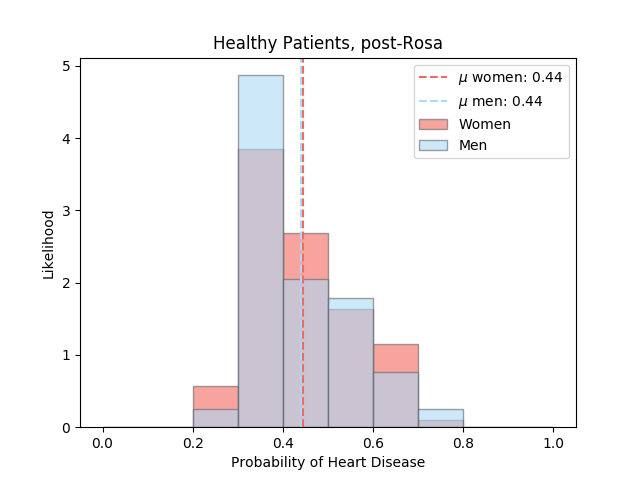} \\
		\includegraphics[width=7cm]{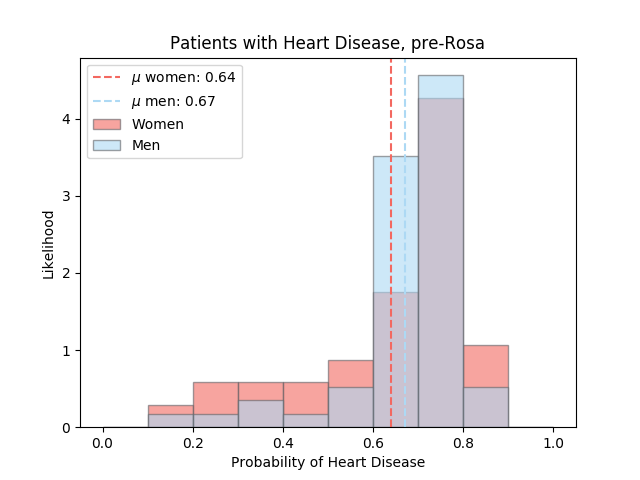} & \includegraphics[width=7cm]{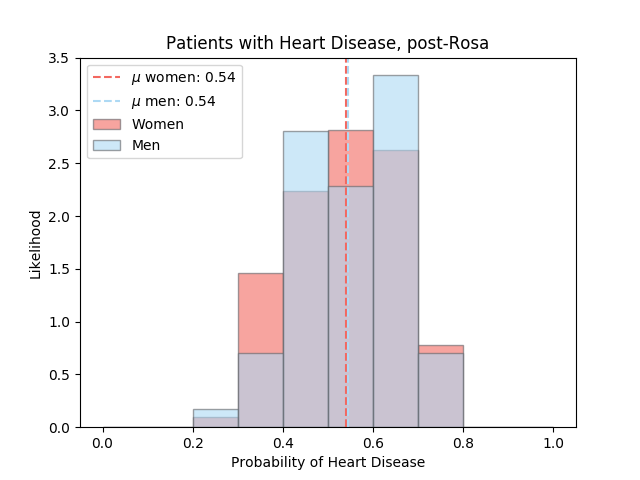} \\
	\end{tabular}
	\caption{Higher model estimates of heart disease were assigned to men compared to women, both for patients with and without heart disease before the data was pre-processed by Rosa. Using the same model with data pre-processed by Rosa, the model assigns equal estimates to men and women with and without heart disease.}
	\label{fig:heart_disease_hist}
\end{figure}

As shown in Figure \ref{fig:heart_disease_hist} and \ref{fig:heart_disease_table}, the model estimates of heart disease for men are clearly higher than for women without data pre-processing with Rosa, regardless of whether the patient had heart disease or not. Using data that has been pre-processed by Rosa, this bias is significantly reduced.

\begin{figure}[H]
	\vspace{20pt}
	    \centering
    \begin{tabular}{r||cc|cc}
         & \multicolumn{2}{c|}{\textbf{Original data}}&\multicolumn{2}{c}{\textbf{Rosa pre-processed}} \\
         &Healthy & Heart Disease & Healthy &Heart Disease\\
         \hline
         \hline
         $\mu$ Women & 0.37 & 0.64 & 0.44 & 0.54\\
         $\mu$ Men & 0.44 & 0.67 & 0.44 & 0.54\\
         $\mu$ Diff & 0.07 & 0.03 & 0.00 & 0.00 \\
         \hline
         $\sigma$ Women & 0.24 & 0.18 & 0.11 & 0.12 \\
         $\sigma$ Men & 0.20 & 0.16 & 0.11 & 0.11 \\
         $\sigma$ Average & 0.22 & 0.16 & 0.11 & 0.11 \\
         \hline
         $\mu$ Diff / $\sigma$ Average & 0.30 & 0.20 & 0.04 & 0.05 \\
    \end{tabular}
    
    \caption{Women received lower estimates of heart disease from our regression model compared to men, independent of whether they had heart disease or not. This means that women are more likely to be incorrectly given a negative diagnosis when they actually have heart disease, compared to men. The amount of bias in the model has been quantified by dividing the difference in mean estimate for men and women with and without heart disease by the average standard deviation of the two distributions. The higher the value the greater the bias. Using Rosa has significantly decreased the level of bias in the estimates both for those with heart disease and those without.}
	\label{fig:heart_disease_table}
\end{figure}

The accuracy of the model in diagnosing heart disease  using biased data was 0.74, and post-Rosa it was 0.67, although it is not reasonable to directly compare the accuracy of a biased model with the accuracy of an unbiased model. This is because predicting certain outcomes with and without bias are fundamentally different tasks.

\subsubsection{Discussion}
We found that a simple model to predict heart disease was biased with respect to gender. This fits well with research on the under-/mis-diagnosis of heart disease in women compared to men \cite{wu2018editor}. The dataset may suffer from two problems which could lead to biased predictions: 1) the dataset is likely to contain an over-representation of men compared to the true population of heart disease sufferers, as women are less likely be diagnosed correctly; and 2) signals in the data which may lead to good predictions for men do not necessarily apply to women.

After de-biasing with Rosa, the predictions made using the heart disease dataset had significantly reduced bias. This has significant implications, as it means that existing data collected from clinical trials on men can be used to predict risk of heart disease without disadvantaging women.

\subsection{Predicting the Economic Needs Index of Schools}
As the financial situation of a school can have a large impact on the success of its students, it is important to identify and direct resources to financially disadvantaged schools.
In New York City, elite schools admit students based on the results of a single test: the Specialized High Schools Admissions Test, or SHSAT. Unfortunately, due to the lack of resources available to schools in deprived areas, and also the link between race and socio-economic status, there is little diversity in those admitted to these elite high schools. This is a problem that replicates across many cities and many countries.

In order to counter this problem, more resources must be directed to under-performing schools, allowing students from deprived areas to catch-up with their less deprived counterparts. 

A key variable indicating whether students at a particular school are likely to benefit from additional resources is the economic needs index of the school. The economic needs index of the school is the average economic needs index of its pupils, and is based on factors such as whether the student is eligible for assistance from the state and whether they have lived in temporary housing \cite{economic_needs}. 

It is important that the economic needs index can be estimated accurately, however, as race correlates with socio-economic status, there is opportunity for inadvertent racial discrimination in estimates (as we will demonstrate).

\subsubsection{Data and Methods}

PASSNYC is a not-for-profit organisation that uses public data to identify promising students in New York's under-performing schools. They direct resources to these schools in order to increase the diversity of students taking SHSAT.

We used a dataset on schools in New York City, compiled by PASSNYC, to predict the economic needs index of students. The \emph{PASSNYC dataset} is available at \href{https://www.kaggle.com/passnyc/data-science-for-good}{Kaggle.} 

We converted the `Percent Black / Hispanic' column to a binary variable which indicated whether the school had a majority of black students or not, in order to ease the identification of bias.

We used the Ridge regression model from Python's sklearn package to predict the economic needs index of each school while excluding the race variable from the data, and looked for racial bias in the predictions. We then debiased the data with respect to race, and made another set of predictions using the debiased data to see whether the bias had decreased when the data was pre-processed using Rosa.

\subsubsection{Results}


\begin{figure}[H]
	\centering
	\begin{tabular}{ll}
		\includegraphics[width=7cm]{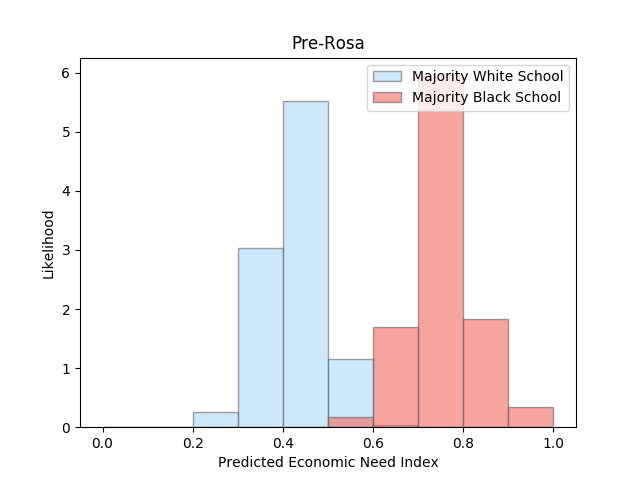} & \includegraphics[width=7cm]{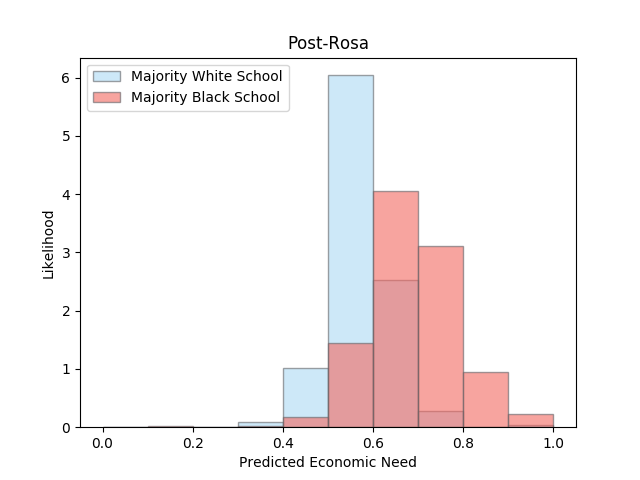} \\
	\end{tabular}
	\caption{Before data  pre-processing by Rosa, schools with a majority of black students receive much higher estimates of economic needs index than schools with a majority of white students, with very little overlap. Using the same model with data pre-processed by Rosa, the estimates are far less polarised.}
	\label{fig:passnyc_hist}
\end{figure}

\begin{figure}[H]
	\vspace{20pt}
	    \centering
    \begin{tabular}{r||c|cc}
        & True Values & Pre-Rosa & Post-Rosa \\
         \hline
         \hline
         $\mu$ Majority Black & 0.76 & 0.75 & 0.69 \\
         $\mu$ Majority White & 0.42 & 0.44 & 0.57 \\
         $\mu$ Diff & 0.34 & 0.31 & 0.12 \\
         \hline
         $\sigma$ Majority Black & 0.13 & 0.08 & 0.11 \\
         $\sigma$ Majority White & 0.19 & 0.06 &0.08 \\
         $\sigma$ Average & 0.16 & 0.07 & 0.095 \\
         \hline
         $\mu$ Diff / $\sigma$ Average & 1.93 & 4.43 & 1.26 \\
    \end{tabular}
    
    \caption{The amount of bias in the model has been quantified by dividing the difference in mean estimate for majority black and majority white schools by the average standard deviation of the two distributions (standardised difference). Prior to data de-biasing using Rosa, the model significantly overestimates the true standardised difference between the economic needs index of majority black schools and majority white schools. After pre-processing with Rosa, this standardised difference is much closer to the true value, with a slight underestimation.}
	\label{fig:passnyc_table}
\end{figure}

 The $R^2$ for predictions prior to debiasing was 0.57, and post-debiasing it was 0.35, although we cannot properly assess model accuracy on biased data.

\subsubsection{Discussion}

We found that without using Rosa our model overestimated the economic needs index of majority black schools, and underestimated the economic needs of majority white schools. This is problematic as it means that white students from disadvantaged backgrounds may not receive the same level of support as similarly disadvantaged black students.

After debiasing the dataset with Rosa, the difference in standardised mean economic needs index estimate assigned by our model was much closer to the true value than before de-biasing. This means that organisations like PASSNYC can better allocate resources by predicting the economic need of students without racial bias.

\subsection{Communities and Crime}
\label{sec:communities}

Being able to predict the rate of crime in different communities is of interest to law enforcement bodies, as it can allow them to better distribute resources such as police officers.

However, there is a significant body of research to suggest that Black people are unfairly represented in criminal datasets due to the racial bias of those responsible for enacting the law \cite{ dabney2004actually, schreer2009shopping, shallice1990black}. In certain circumstances, it has even been found that the true rate of crime for Black persons is likely equal to that of White persons, despite large differences in the data collected by law enforcement bodies \cite{bunting2013war, piquero2008assessing}.

This is highly problematic, as any model used to predict crime rate based on such datasets is likely to overestimate the strength of the true relationship between race and crime, leading to biased predictions. In this example it might lead to an unnecessary direction of police resources to communities with a large Black population. In turn, this is likely to lead to a greater rate of arrest of Black individuals (as there will be more police officers in areas with a large Black population), further strengthening the bias in crime datasets.

\subsubsection{Data and Methods}
We used a dataset on communities and crime within the United States (combining data from the 1990 US Census, law enforcement data from the 1990 US LEMAS survey, and crime data from the 1995 FBI UCR) to predict the rate of violent crime in different communities. 

The \emph{Communities and Crime dataset} is available from the \href{https://archive.ics.uci.edu/ml/datasets/Communities+and+Crime}{UCI Machine Learning Repository}. We discarded 22 of the most sparsely populated columns, and converted the `RacePercentBlack' column to a binary label indicating whether the black proportion in a given community is in the upper quartile across all communities in the dataset. We removed all other columns that contained information about the racial profile of each community.

We used the Linear Regression model in Python's sklearn package to predict the rate of violent crime in each community while excluding the race variable from the data, and looked for bias with respect to race. We then debiased the data with respect to race, and made another set of predictions using the exact same DA pipeline to see whether the bias had truly been removed.

\subsubsection{Results}

\begin{figure}[H]
	\centering
	\begin{tabular}{cc}
		\includegraphics[width=7cm]{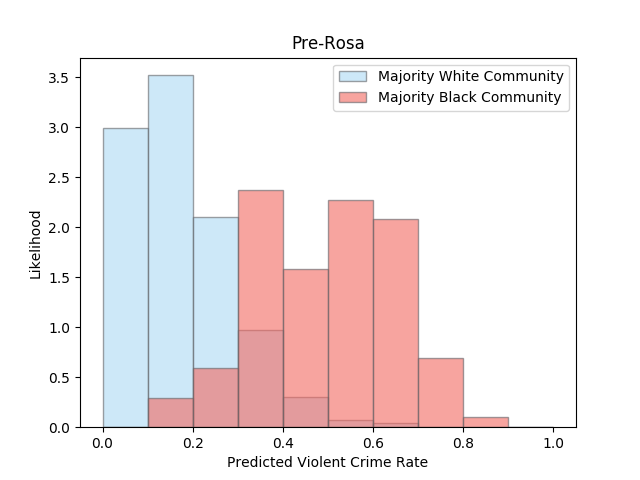} & \includegraphics[width=7cm]{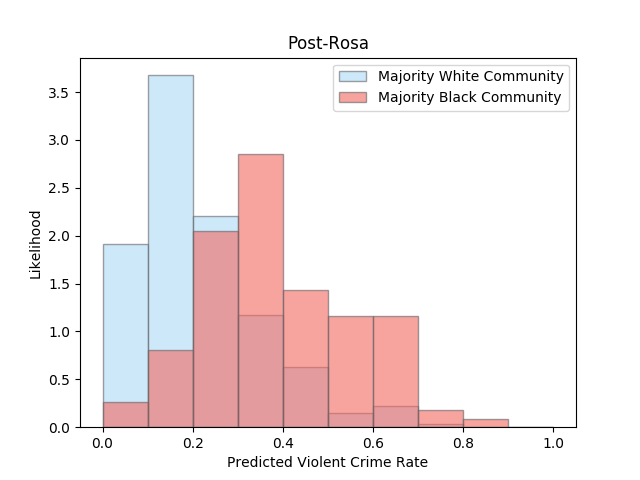} \\
	\end{tabular}
	\caption{Before the data was pre-processed by Rosa, our model assigns much higher estimates of violent crime rate to majority Black communities compared to majority White communities. After the data has been pre-processed by Rosa, there is less discrepancy between model estimates for majority Black and majority White communities. }
	\label{fig:communitites_predictions}
\end{figure}

\begin{figure}[H]
	\vspace{20pt}
	    \centering
    \begin{tabular}{r||c|cc}
        & True Values & Pre-Rosa & Post-Rosa \\
         \hline
         \hline
         $\mu$ Black Upper Quart & 0.49 & 0.51 & 0.39 \\
         $\mu$ Black Lower Quarts & 0.16 & 0.15 & 0.20 \\
         $\mu$ Diff & 0.33 & 0.36 & 0.19 \\
         \hline
         $\sigma$ Black Upper Quart & 0.27 & 0.17 & 0.29 \\
         $\sigma$ White Lower Quarts & 0.16 & 0.13 & 0.17 \\
         $\sigma$ Average & 0.215 & 0.15 & 0.23 \\
         \hline
         $\mu$ Diff / $\sigma$ Average & 1.53 & 2.40 & 0.83 \\
    \end{tabular}
    \caption{There is a large difference in the rate of violent crime between majority Black and majority White communities in the \emph{Communities and Crime dataset}, despite evidence to suggest that the true difference should be minimal. This is reflected in the estimates from the model before data pre-processing with Rosa. The amount of bias in the model has been quantified by dividing the difference in mean estimate for upper-quartile black and lower three-quartile black communities by the average standard deviation of the two distributions. The higher the value the greater the bias. The pre-Rosa model has even greater bias than the original dataset. After data pre-processing with Rosa, our model estimates have less bias than the original dataset.}
	\label{fig:communities_table}
\end{figure}

 The $R^2$ for predictions prior to debiasing was 0.69, and post-debiasing it was 0.52, although we cannot properly assess model accuracy on biased data.

\subsubsection{Discussion}

There is evidence to suggest that there is only a small difference in the rate of crime for Black persons and White populations, however, most criminal datasets currently have a large over-representation of Black persons due to the well documented biases of those enacting the law. This is problematic, as police resources may be incorrectly distributed based on this data, further strengthening the existing bias.

We found that our simple model overestimated violent crime rate in communities with a high proportion of Black residents. After pre-processing the data using Rosa, there was much less difference in the model estimates for majority Black and majority White communities.

Although the difference in model estimates for communities with a large proportion of Black residents and those with a large proportion of White residents after pre-processing with Rosa was smaller than the true difference in the dataset, extensive research suggests the dataset itself is biased, and therefore it cannot be used as a benchmark. We must instead aim for a level of bias below the true dataset bias in order to prevent the further propagation of racial bias through the criminal justice system. This was achieved by using Rosa.

\section{Discussion}

Rosa is a free, web-based tool, which uses the principles of Fair Adversarial Networks [FANs] to remove bias from a dataset with respect to a certain characteristic or characteristics. In this paper we demonstrated the bias removing capabilities of Rosa on five datasets which were related to real and current issues in the world of Data Analytics [DA] and Machine Learning [ML]. These datasets contained racial, gender or age bias that made the results of our analysis clearly biased as well; however the bias was successfully removed/significantly decreased each time we have used Rosa as a pre-processing step in our analysis.

The main advantage of Rosa is its wide applicability and simplicity for the end user. Rosa is stand-alone and does not require any integration into the  ML pipeline that a dataset is intended for. It is therefore compatible with a wide range of ML and data analysis techniques. Less technically minded users may choose to use a graphical user interface such as the one available at \href{https://rosa.illumr.com}{rosa.illumr.com} to pre-process their data for further use in software such as MS Excel or IBM SPSS. A data scientist comfortable with scripting, on the other hand, can use the Rosa API directly. In such cases debiasing data becomes a single line of code.

It should be noted that although we have demonstrated the efficacy of Rosa at removing bias with respect to protected characteristics \cite{act2010c}, Rosa can remove any type of bias. For example, it might be desirable to remove regional bias when evaluating performance of an enterprise's regional offices. In such cases performance of employees in different parts of the country may not be comparable without removing the bias of the region.


There were slight decreases in prediction accuracy across all models after data debiasing. However, because the datasets themselves were all biased in some respect, we cannot directly compare model accuracy before and after debiasing. The model trained on the biased dataset was only tested on biased data - we do not know what its performance would be on fair data. Tests on synthetic datasets with artificially injected bias indicate that the accuracy increases when using Rosa (running the DA pipeline with biased data but evaluating the performance on data before artificial biasing). This result is, however, reliant on the assumptions about how human biases work and therefore do not provide a definite proof of increase in accuracy when using Rosa. The nature of the problem that Rosa addresses makes it impossible to prove an increase in accuracy in the real-world.

The stochastic nature of both the simple DA models and the FANs behind Rosa mean that the results of similar experiments will vary every time, and for a thorough analysis should be repeated multiple times. In our analyses for this paper, we present a randomly selected run of the DA pipeline to keep the matter simple. We have observed only very marginal variance in the output with repeated runs.

As demonstrated in this paper, Rosa is capable of removing many different types of bias, with no change in approach taken by the end user other than selecting the right characteristic. This ease-of-use is unrivalled by alternative bias removal methods.

\bibliographystyle{plain}
\bibliography{mybib}{}
	
\end{document}